\documentclass[a4paper,fleqn]{cas-sc}

\usepackage[authoryear]{natbib}
\usepackage{mathtools}

\begin{document}
\let\WriteBookmarks\relax
\def\floatpagepagefraction{1}
\def\textpagefraction{.001}
\shorttitle{Combining data assimilation and machine learning}
\shortauthors{J. Brajard et~al.}

\title [mode = title]{Combining data assimilation and machine learning to emulate a dynamical model from sparse and noisy observations: a case study with the Lorenz 96 model}

\author[1,2]{Julien Brajard}[orcid=0000-0003-0634-1482]
\cormark[1]
\ead{julien.brajard@nersc.no, julien.brajard@sorbonne-universite.fr}
\ead[url]{https://jbrlod.locean-ipsl.upmc.fr/website/index.html}

\credit{Conceptualization, Methodology, Software, Validation, Writing - Original Draft}

\author[3,4]{Alberto Carrassi}[orcid=0000-0003-0722-5600]

\credit{Methodology, Validation, Writing - Review \& Editing, Project administration, Funding acquisition}

\author[5]{Marc Bocquet}[orcid=0000-0003-2675-0347]

\credit{Methodology, Validation, Writing - Review \& Editing}

\author[1]{Laurent Bertino}[orcid=0000-0002-1220-7207]

\credit{Methodology, Validation, Writing - Review \& Editing,  Project administration, Funding acquisition}

\address[1]{Nansen Center, Thorm{\o}hlensgate 47, 5006, Bergen, Norway}
\address[2]{Sorbonne  University, CNRS-IRD-MNHN, LOCEAN, Paris, France}
\address[3]{Departement of Meteorology, University of Reading and NCEO, United-Kingdom }
\address[4]{Mathematical Institute - University of Utrecht}
\address[5]{CEREA, joint laboratory \'Ecole des Ponts ParisTech and EDF R\&D, Universit\'e Paris-Est, Champs-sur-Marne, France}

\cortext[cor1]{Corresponding author}



\begin{abstract}
A novel method, based on the combination of data assimilation and machine learning is introduced. The new hybrid approach is designed for a two-fold scope: (i) emulating hidden, possibly chaotic, dynamics and (ii) predicting their future states.
The method consists in applying iteratively a data assimilation step, here an ensemble Kalman filter, and a neural network. Data assimilation is used to optimally combine a surrogate model with sparse noisy data. The output analysis is spatially complete and is used as a training set by the neural network to update the surrogate model. The two steps are then repeated iteratively. Numerical experiments have been carried out using the chaotic 40-variables Lorenz 96 model, proving both convergence and statistical skill of the proposed hybrid approach. 
The surrogate model shows short-term forecast skill up to two Lyapunov times, the retrieval of positive Lyapunov exponents as well as the more energetic frequencies of the power density spectrum. The sensitivity of the method to critical setup parameters is also presented: the forecast skill decreases smoothly with increased observational noise but drops abruptly if less than half of the model domain is observed. 
The successful synergy between data assimilation and machine learning, proven here with a low-dimensional system, encourages further investigation of such hybrids with more sophisticated dynamics. 
\end{abstract}

\begin{highlights}
\item We address the problem of dynamics emulation from sparse and noisy observations.
\item An algorithm combining data assimilation and machine learning is applied.
\item The approach is tested on the chaotic 40-variables Lorenz 96 model.
\item The output of the algorithm is a data-driven surrogate numerical model.
\item The surrogate model is validated on both forecast skill and long-term properties.
\end{highlights}

\begin{keywords}
Data assimilation \sep Machine Learning \sep Emulator \sep Dynamical model \sep Observations
\end{keywords}

\maketitle

\section{Introduction}
Geophysical fluid dynamics is a domain of science in which the physical and dynamical laws governing the systems are reasonably well known. This knowledge is expressed by partial differential equations that are then discretised into numerical models~\citep{Randall2007ClimateEvaluation}.
However, due to the misrepresentation of unresolved small-scales features or to neglected physical processes, parts of the numerical models have to be represented by empirical sub-models or parameterizations. Earth observations are thus needed for two tasks: first for model tuning and selection~\citep[best realism of the model; i.e. estimating/selecting the best possible parameterizations; see e.g.][]{metref2019estimating} and then for data assimilation ~\citep[best accuracy of the model; i.e. estimating the system state; see e.g.][]{Carrassi2018DataPerspectives}. Sometimes the same data are used for both tasks. 
In the last decades, the volume and quality of Earth observations have increased dramatically, particularly thanks to remote sensing~\citep[see, e.g.,][]{Kuenzer2014EarthBottlenecks}. 
At the same time, new developments in machine learning (ML), particularly deep learning~\citep{Lecun2015DeepLearning}, have demonstrated impressive skills in reproducing complex spatiotemporal processes~\citep[see e.g.][for video processing]{Tran2018LearningNetworks} by efficiently using a huge amount of data, thus paving the path for their use in Earth system science~\citep{Reichstein2019DeepScience}.

Forecasting and simulating are two different objectives that can both be achieved using ML.
Various ML algorithms have been already applied to produce surrogate models of fully observed low-order chaotic systems and then used for forecasting purposes; examples include reservoir modeling~\citep{Pathak2017UsingData, Pathak2018Model-FreeApproach}, residual neural network~\citep{fablet2018} and recurrent networks~\citep{Park2002BilinearNetwork,Park2010ANetwork}. Machine learning has also been applied for nowcasting based on real observations, such as sea surface temperature~\citep{DeBezenac2017TowardsModeling} or precipitation  ~\citep{Shi2015ConvolutionalNowcasting,Shi2017DeepModel}. Along with these advancements in ML, the issue of determining a surrogate model of an unknown underlying process based on observations has also been addressed using sparse regression~\citep{Brunton2017ChaosSystem, zhang2018robust} and, more recently, data assimilation~\citep{Bocquet2019DataModels}.

Diverse numerical results have proven the effectiveness of ML to reconstruct dynamics under the condition that the ML algorithm is trained on noise-free and complete observations of the system (i.e. when the system state is fully observed). Nevertheless, these circumstances are almost never encountered in practice and certainly not in the geosciences.  In the case of partial and noisy observations, one can use a nearest-neighbour or an analogue approach: it consists in finding similar past data (if available) as a forecast~\citep{Lguensat2017}. 
Machine learning techniques have been also applied in situations where only a dense portion of the system is observed~\citep[e.g.][]{DeBezenac2017TowardsModeling,Lu2017ReservoirSystems} or when the observations are sub-sampled in time~\citep{Nguyen2019EM-likeObservations}.

Most of the ML algorithms used in the aforementioned studies are not suited to realistic cases of noisy and sparse observations. By ``sparse'', we mean here that the state of the system is not densely observed, and furthermore that the locations and the number of these sparse observations may vary in space and time. 

Data assimilation (DA) is a natural framework to deal with this observational scenario: it is designed to estimate the state of a system given noisy, unevenly distributed, observations and a dynamical model~\citep{Carrassi2018DataPerspectives}. The output of DA depends explicitly on the uncertainties in the numerical model~\citep{Harlim2017ModelAssimilation}, which has led to developing techniques for taking into account model errors in the DA process. This can be done by a parametrization of the model error embedded in the model equations~\citep{Aster2005ParameterGeophysics,Carrassi2011StateDynamics,Bocquet2012} or in form of a stochastic noise added to the deterministic model~\citep[e.g.][]{Tremolet2006Accounting4D-Var,Ruiz2013EstimatingReview, Raanes2015ExtendingMethods, Sakov2018AnError}. In any case, a dynamical model must be at disposal and its presence and use are key in DA. In~\citet{Bocquet2019DataModels} though, this constraint is relaxed to the point that only the general form of the differential equations, representing the model, is specified. Notably, and a key to the present study, it was shown that the optimization problem solved by DA in the presence of a model error is equivalent to a ML problem. A similar equivalence was also shown by~\citet{Abarbanel2018MachineProblems}, starting from the point of view of a ML problem. A pure DA approach, however, does not leverage on recent ML developments, that bring flexibility and facilitate parallel  calculations. By including explicit or implicit regularization processes, ML algorithms make possible optimizing in high-dimension without the need for additional information under the form of an explicit prior.

This paper stands at the crossroads of DA and ML, and focuses on the cases where both the system state and its dynamical model have to infer based on noisy and sparse observations. The proposed hybrid algorithm relies on DA to estimate the state of the system and on ML to emulate the surrogate model. 

The paper is organised as follows. In section~\ref{sec:method}, we formulate the problem and detail how DA and ML are combined. In section~\ref{sec:experiments}, we present the experimental setup whereas section~\ref{sec:results} describes the numerical results using different metrics and discusses the algorithm sensitivity to the number of observations and their noise statistics as well as to  other control parameters. Conclusions and perspective are drawn in section~\ref{sec:conclusion}.

\section{\label{sec:method}Methodology}
\subsection{Definition of the problem}
Let us consider a time series of  multi-dimensional observations $ \mathbf{y}^\textrm{obs}_k \in  \mathbb{R}^p$ of an unknown process $\mathbf{x}_k\in \mathbb{R}^m$:
\begin{equation}
    \mathbf{y}^\textrm{obs}_k = \mathcal{H}_k (\mathbf{x}_k) + \boldsymbol{\epsilon}^{\textrm{obs}}_k,
    \label{eq:obs}
\end{equation}
where $0 \leq k \leq K$ is the index corresponding to the observation time $t_k$, and $\mathcal{H}_k: \mathbb{R}^m \rightarrow  \mathbb{R}^p$ is the observation operator (supposed to be known). The observation error $\boldsymbol{\epsilon}^{\textrm{obs}}_k$ is assumed to follow a normal distribution of mean zero and covariance matrix $\mathbf{R}_k$.

In the following, we will assume for the sake of simplicity that the
number of observations, $p$, and their noise level does not change with time. Furthermore, the observations are assumed to be spatially uncorrelated, implying that $\mathbf{R}$ is diagonal such that
\begin{equation}
  \mathbf{R}_k \equiv (\sigma^\textrm{obs})^2 \mathbf{I}_p,
  \label{eq:sigmaobs}
 \end{equation}
  where $\sigma^\textrm{obs}$ is the standard deviation of the observation error and $\mathbf{I}_p$ is the identity matrix of size $p \times p$. We will also consider a regular time discretisation step such as: $t_{k+1}-t_k = h$ for all $k$.

We suppose that $\mathbf{x}_k$ is a time-discretisation of a continuous process $\mathbf{x}$, which obeys to an unknown ordinary differential equation of the form
\begin{equation}
  \frac{\textrm{d}\mathbf{x}}{\textrm{d}t} = \mathcal{M}(\mathbf{x}).
\label{eq:dynmodel}
\end{equation}

Our goal is to predict the state at time $t_{k+1}$ given the state at time $t_k$, which consists in deriving a surrogate model $\mathcal{G}$ of the so-called resolvent of $\mathcal{M}$ between $t_k$ and $t_{k+1}$ 
defined as:
\begin{equation}
    \mathbf{x}_{k+1} = \mathcal{G}(\mathbf{x}_k)  + \boldsymbol{\epsilon}^\textrm{m}_k =  \mathbf{x}_k +\smashoperator{\int_{t_k}^{t_{k+1}}}\mathcal{M}(\mathbf{x}) \, \textrm{d}t, \label{eq:resolvent}
\end{equation}
where $\boldsymbol{\epsilon}^\textrm{m}_k$ is the error of the surrogate model $\mathcal{G}$.

\subsection{Convolutional neural network as surrogate model}
As stated in the introduction, several papers have used convolutional neural networks for representing surrogate models~\citep[see, e.g.,][]{Shi2015ConvolutionalNowcasting, DeBezenac2017TowardsModeling,fablet2018}. 
Equation~(\ref{eq:resolvent}) being in the incremental form $\mathbf{x}_{k+1}=\mathbf{x}_k + \cdots$, one-block residual networks are suitable~\citep{He2015DeepRecognition}. 
So, our neural network can be expressed as a parametric function $\mathcal{G}_\mathbf{W}(\mathbf{x})$:
\begin{equation}
\mathcal{G}_\mathbf{W}(\mathbf{x}_k) = \mathbf{x}_k + f_\textrm{nn}(\mathbf{x}_k,\mathbf{W}),
\end{equation}
where $f_\textrm{nn}$ is a neural network and $\mathbf{W}$ its weights; $f_\textrm{nn}$ is composed of convolutive layers~\citep{Goodfellow-2016-book}. Convolutive layers apply a convolution acting in a local spatial neigbourhood around each state variable $x_{n,k}$ of the field $\mathbf{x}_k = \left[x_{0,k}, \cdots, x_{m-1,k}\right]$. This is equivalent to the locality hypothesis as defined in, e.g.~\citet{Bocquet2019DataModels}. It is the underlying hypothesis to the use of localization in ensemble DA, a regularization process that discards spurious long-range correlations caused by a limited number of members in the ensemble~\citep{evensen2003ensemble}. Note that it does not prevent long-distance correlations arising from the time integration.

The determination of the optimal weights $\mathbf{W}$ is achieved via an iterative minimization process of a loss function: this process is referred to as the \textit{training phase}. The loss function reads:
\begin{equation}
L(\mathbf{W}) = \sum_{k=0}^{K-N_\textrm{f}-1} \sum_{i=1}^{N_\textrm{f}}\left\lVert
\mathcal{G}_\mathbf{W}^{(i)}(\mathbf{x}_k) - \mathbf{x}_{k+i}
\right\rVert^2_{\mathbf{P}_k^{-1}}, \label{eq:loss}
\end{equation}
where $N_\textrm{f}$ is the number of time steps corresponding to the forecast lead time on which the error between the simulation and the target is minimized;  $\mathbf{P}_k$ is a symmetric, semi-definite positive matrix defining the norm $\lVert \mathbf{x} \rVert  ^2_{\mathbf{P}_k^{-1}} = \mathbf{x}^\textrm{T}\mathbf{P}_k^{-1}\mathbf{x}$. Note that $\mathbf{P}_k$ plays the role of the surrogate model error covariance matrix. The introduction of this time-dependent matrix aims at giving different weights to each state during the optimization process, depending on how uncertain they are: an uncertain state has a low weight. We will clarify in the next sections how  $\mathbf{P}_k$ is itself estimated using DA. 

The formalism used here is close to what is proposed in~\citet{ E2017ASystems, Chang2017Multi-levelView, fablet2018} in which the neural network is identified with a dynamical model expressed as Ordinary Differential Equations (ODEs). One major difference is that the algorithm does not aim here at identifying the function $\mathcal{M}$ of an ODE, see Eq.~(\ref{eq:dynmodel}), but at reproducing a resolvent of the underlying dynamics, see Eq.~(\ref{eq:resolvent}).
We also highlight that a prerequisite to train this neural network is to have access to a time series of the complete state field $\mathbf{x}_{1:K}$ for which we will rely upon DA.

\subsection{Data assimilation}
In this work, we use the finite-size ensemble  Kalman filter~\citep{Bocquet2011EnsembleInflation,Bocquet2015ExpandingInflation}, hereafter denoted EnKF-N. However, our approach is not tied to any particular DA algorithm and is straightforwardly applicable to any adequate DA method for the problem at hand. For example, a smoother would lead to more accurate results but would be more costly and is thus less common in the operational DA community. Our choice of the EnKF-N is motivated by its efficiency, its high accuracy for low-dimensional systems, and its implicit estimation of the inflation factor that would otherwise have had to be tuned.

The EnKF-N is a sequential ensemble DA technique.
The forecast matrix at time $t_k$ is defined by

$\mathbf{X}_k^{\textrm{f}} \equiv \left [ \mathbf{x}_{k,1}^{\textrm{f}}, \cdots,   \mathbf{x}_{k,p}^{\textrm{f}}, \cdots,  \mathbf{x}_{k,N}^{\textrm{f}} \right]\in  \mathbb{R}^{m \times N}$.

Similarly, the analysis matrix at time $t_k$ is defined by

$\mathbf{X}_k^{\textrm{a}} \equiv \left [ \mathbf{x}_{k,1}^{\textrm{a}}, \cdots,   \mathbf{x}_{k,p}^{\textrm{a}}, \cdots,  \mathbf{x}_{k,N}^{\textrm{a}} \right]\in  \mathbb{R}^{m \times N}$.

The members  $\mathbf{x}_{k,p}^\textrm{f}$ are computed recursively in the forecast step:
\begin{equation}
    \mathbf{x}_{k,p}^\textrm{f}= \mathcal{G}(\mathbf{x}_{k-1,p}^\textrm{a}) + \boldsymbol{\epsilon}_{k,p}^\textrm{m}, \label{eq:forecast}
\end{equation}
where $\boldsymbol{\epsilon}_{k,p}^\textrm{m}$ is the model noise for member $p$ at time $t_k$.

Given the observations  $\mathbf{y}^\textrm{obs}_k$ and the forecast ensemble $\mathbf{X}_k^\textrm{f}$, the analysis step yields an updated ensemble $\mathbf{X}_k^\textrm{a}$ following the algorithm 3.4 in~\citet{Bocquet2011EnsembleInflation}. The mean $\mathbf{x}^\textrm{a}_k$ of the ensemble state and the error covariance matrix $\mathbf{P}^\textrm{a}_k$ are estimated by
\begin{align}
& \mathbf{x}^\textrm{a}_k = \frac{1}{N}\sum_{p=1}^N\mathbf{x}_{k,p}^\textrm{a},  \label{eq:analysis}\\
& \mathbf{P}^\textrm{a}_k = \frac{1}{N-1}(\mathbf{X}_k^\textrm{a}-\mathbf{x}^\textrm{a}_k\mathbf{1}_N)(\mathbf{X}_k^\textrm{a}-\mathbf{x}^\textrm{a}_k\mathbf{1}_N)^\textrm{T}, \label{eq:covar}
\end{align}
where $\mathbf{1}_N$ is the row vector composed of N ones.

\subsection{Combining data assimilation and machine learning}
The general idea of the proposed algorithm is to combine DA and ML: the neural network provides a surrogate forward model to DA and, reciprocally, DA provides a time series of complete states to train the neural network.
\begin{figure}
  \centering
\includegraphics[scale=.75]{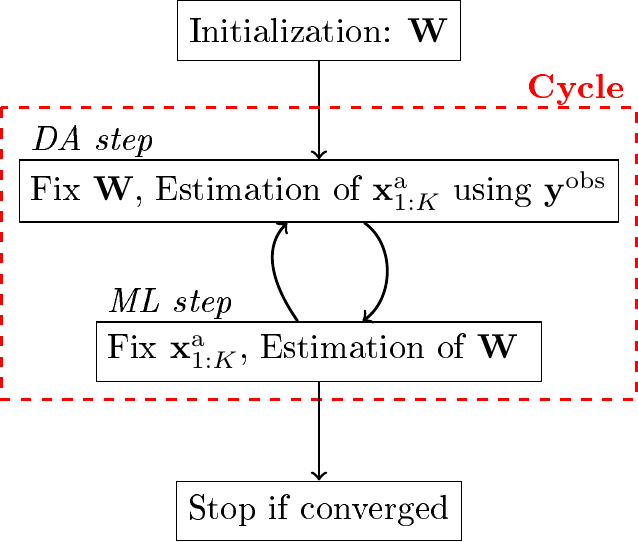}
\caption{\label{fig:algo} Scheme of the two-step procedure algorithm. The first step, DA (data assimilation), estimates the state field $\mathbf{x}_{1:K}$ based on the fixed model parameters $\mathbf{W}$. The second step, ML (machine learning), estimates the parameters $\mathbf{W}$ of the neural network-based model by using the state field $\mathbf{x}_{1:K}$. $\mathbf{P}_k$, defined in Eq.~(\ref{eq:covar}), is passed from the DA to the ML step.}
\end{figure}
An illustration of the algorithm is displayed in Fig.~\ref{fig:algo}.
At each cycle, a DA step feeds into a ML step and vice-versa. The DA step is applied to the whole observation time series according to Eq.~(\ref{eq:forecast}-\ref{eq:covar}) using the surrogate model $\mathcal{G}_W$ as the forecast model. 
The ML step is done by minimizing the loss in Eq.~(\ref{eq:loss}). 
In this work, the matrix $\mathbf{P}_k$ used in the norm of the loss $L(\mathbf{W})$ is the diagonal of the matrix defined in Eq.~(\ref{eq:covar}). 
This simplification is done to gain computational efficiency and to reduce numerical bias potentially arising from a biased estimation of the correlation terms in the covariance matrix. 
However besides this simplification, it is worth noticing that, being based on the ensemble, 
the covariance matrix $\mathbf{P}^\textrm{a}_k$ is time-dependent like the analysis error in the state estimate.

Two particular features of the proposed algorithm can be highlighted here:
\begin{itemize}
    \item The algorithm leverages on ML libraries to estimate the parameters of the surrogate model $\mathcal{G}_W$. It can thus benefit from the high speedups, parallel computing and regularization processes of these libraries enabling optimization of high-dimensional $\mathbf{W}\!$. 
    \item The two steps in each cycle are separate, making the algorithm very flexible. The choices of the ML and the DA algorithms are independent. They can be changed or combined with an external system. 
\end{itemize}

The initialization step is critical as the convergence of the hybrid algorithm is not guaranteed for any initial choice of $\mathbf{W}$. For simple cases, it is possible to initialize the weights of the neural network randomly. In the following, we will use an interpolated field to train the initial weights of the neural network.

We also found that the convergence of the algorithm is sensitive to hyperparameters such as the model noise level, Eq.~(\ref{eq:forecast}) and the length $K$ of the observational period. These aspects are discussed in section~\ref{sec:tuning}.

\section{\label{sec:experiments}Numerical experiment setup}

\subsection{\label{sec:setup}Model Setup}
Our combined DA-ML method is tested using the 40-variables Lorenz model~\citep{Lorenz1998OptimalModel} hereafter denoted L96 which is used to produce the synthetic observations. In this idealized case, the surrogate model can be compared with the real underlying dynamic, called the ``true'' model.
The model L96 is defined on a periodic one-dimensional domain by the following set of ordinary differential equations:
\begin{equation}
\frac{\textrm{d}x_n}{\textrm{d}t} = (x_{n+1} - x_{n-2})x_{n-1} - x_n + F,
    \label{eq:L96}
\end{equation}
where $x_n$ ($0 \leq n <m$) is the scalar state variable, $x_m = x_0$, $x_{-1}=x_{m-1}$, $x_{-2}=x_{m-2}$, $m=40$ and $F=8$. The model is integrated using a fourth order Runge-Kutta scheme with a time step $h=0.05$. The resolvent of the true model is denoted $\mathcal{G}_{L96}$.
The model L96, with the values of $m$ and $F$ given above, is chaotic, with the largest Lyapunov exponent $\Lambda_1 \approx 1.67$. In the following, we will use the Lyapunov time unit  $t_\Lambda = \Lambda_1t$ where $t$ is the time in model unit: one Lyapunov time unit corresponds to the time for the error to grow by a factor $e$.

Unless stated otherwise, results will be shown on the so-called ``reference setup'' specified in the following. 
The model is integrated over $K=40,000$ time steps to produce a state vector time series $\mathbf{x}_{1:K}$ considered as the truth. The sensitivity to the dataset length $K$ is explored in section.~\ref{sec:tuning-K}.
In this setup, the observation operator $\mathcal{H}_k$ is defined as a sub-sampling operator that draws randomly $p=20$ values at each time step (corresponding to 50\% of the field) from a uniform distribution changing the observation locations at each time step. Note that variables have a non-zero probability to be observed at any time.  In the more challenging cases where portions of the field are never observed (e.g. fixed observing network, hidden parameters, ...), the setup may have to be adapted to account for an unobserved latent subspace, as proposed for instance in~\citet{ayed2019learning}. 

The observation interval is the same as the integration time step of the model. In a real application, the integration time of the underlying model would have been unknown (the underlying model might not even be specified), so the only known time step is that of the observations~\citep[see the discussion in][]{Bocquet2019DataModels}. The standard deviation of the observational error in Eq.~(\ref{eq:obs},~\ref{eq:sigmaobs}) is set to $\sigma^\textrm{obs}=1$ (about $5\%$ of the total range). The value was chosen sufficiently large so that reducing the noise has a real impact on the results.

\subsection{\label{sec:scores}Scores}
We will assess our methods by several scores, responding to various desirable aspects of the retrieved surrogate model. In particular, our scores assess the forecasting skill (i.e. short-to-medium range) as well as the climate skill (i.e. long-range, statistics). In the context of a specific application, it would be possible to focus exclusively on one aspect.

Given the resolvent of the true model $\mathcal{G}_{L96}$ and the true field $\mathbf{x}_{1:K}$, we compute the following scores:
\begin{itemize}
    \item RMSE-a: aims at assessing the accuracy of the DA step at the end of the procedure using the spatiotemporal root mean square error (RMSE) of the analysis:
    \begin{equation}
        \text{RMSE-a} = 
        \sqrt{\frac{1}{m(K-k_0)}\sum_{k=k_0}^K\sum_{n=0}^{m-1}
        (x_{n,k} - x^\textrm{a}_{n,k})^2}, 
    \end{equation}
    where $\mathbf{x}_k = \left[x_{0,k}, \cdots, x_{m-1,k}\right]$ and $k_0>0$ is the first value considered in the evaluation (we have chosen $k_0=100$ corresponding to $t_\Lambda \approx 8$) so as to remove the effect of the initial state.
    
    \item RMSE-f: evaluates the forecast quality of the surrogate model $\mathcal{G}_\mathbf{W}$. We defined a vector of $P=500$ different initial conditions $\mathbf{x}_0^{1:P}$ that are not consecutive in time and are drawn from another simulation of the true model. These initial condition are then used to initialise the surrogate model and the RMSE-f is calculated for a predicting horizon of $t_i-t_0$:
   
\begin{align}
&      \text{RMSE-f}(t_i-t_0) = \nonumber \\
&   \sqrt{\frac{1}{m P}\sum_{p=1}^P\sum_{n=0}^{m-1}
        \left([\mathcal{G}_{L96}^{(i)}(\mathbf{x}^p_{0})]_n - [\mathcal{G}_\mathbf{W}^{(i)}(\mathbf{x}^p_{0})]_n\right)^2}.
        \label{eq:rmse-f}
\end{align}

    \item Average value. Given that the state is homogeneous, the average is computed both spatially and temporally by the expression
    \begin{equation}
        \overline{\mathbf{x}} = \frac{1}{mK}\sum_{k=1}^K\sum_{n=0}^{m-1}x_{n,k},
    \end{equation}
    where $x_{n,k}$ can be computed either by the true model or by the surrogate model over $K$ time steps.
    \item Lyapunov spectrum: assesses the long-term dynamical property of the surrogate model to be compared with that of the true model.
    The Lyapunov spectrum was computed experimentally over an integration of 100,000 time steps to ensure the convergence of the calculation~\citep{legras1996guide}.
    
    \item Power spectral density: describes the "climate" of the model in terms of its frequency content. The power spectral density is computed on a single grid point using the Welch method~\citep{Welch1967ThePeriodograms}: the time series is split into 62 overlapping segments of 512 points each. A power spectrum is computed on each segment convoluted with a Hanning window and all spectra are then averaged to yield a smooth power spectral density.
    
\end{itemize}

\subsection{Surrogate model optimization setup}

\subsubsection{Data assimilation setup}
Following~\citet{Bocquet2011EnsembleInflation}, we work with an ensemble of size $N=30$ which makes localization unnecessary.  Note however that localization is unavoidable when working in high dimensions, and needs careful tuning.
The initial state is drawn from a sequence calculated from a previous run. The model noise $\boldsymbol{\epsilon}^\textrm{m}$ is drawn from a normal distribution with zero mean and standard deviation $\sigma^\textrm{m} = 0.1$. The sensitivity to the level of model noise in the DA step is discussed in section~\ref{sec:tuning}.

\subsubsection{Neural network setup}
\begin{figure}
  \centering
\includegraphics[scale=.75]{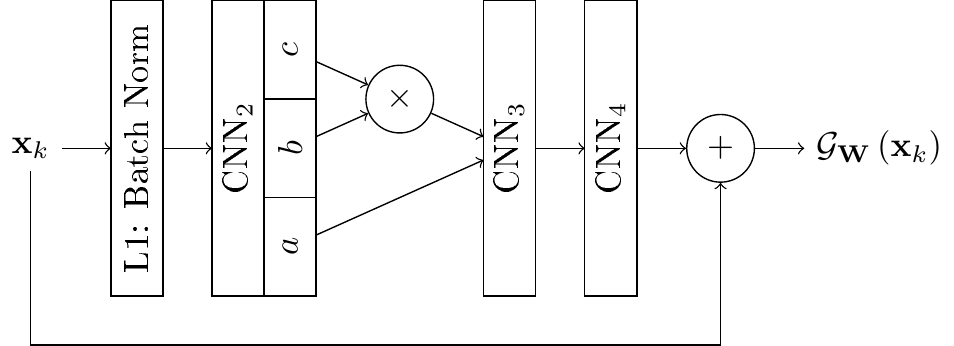}
\caption{\label{fig:nn} Proposed neural network architecture for the surrogate model. The input layer is to the left. }
\end{figure}

The architecture of the neural network (e.g. number/type of layers, activation functions, and others) and the settings of the training (number of epochs, minimizer, batch size) have been determined by cross-validation experiments on two datasets: one with a fully observed field without noise, the other with a cubic interpolation of the observations from the reference setup ($p=20$ observations, and $\sigma^\textrm{obs}=1$). The proposed architecture is shown in Fig.~\ref{fig:nn} whereas its parameters and those of the training phase are detailed in Tab.~\ref{tab:nn}. The bilinear layer in our network aims at facilitating the training when multiplications are involved in the true model. This design is therefore based on a priori knowledge that multiplicative terms are ubiquitous in ODE-based geophysical models. For instance, this is the case of the model described in Eq.~(9) used to illustrate our method. Note that in a similar setup, this architecture is comparable to the one proposed by~\citet{fablet2018}, but given that our goal is not the identification of the model itself, our architecture is not embedded in a time-integration step (here a fourth-order Runge-Kutta scheme). In this way, the neural network is emulating the resolvent of the model but may not represent the model itself.
As the numerical experiments are conducted on a periodic spatial domain, (see section~\ref{sec:setup}), the input of the neural network is 1-D periodic at the boundaries too.

\begin{table}
\caption{\label{tab:nn}Setup of the architecture (cf Fig.~\ref{fig:nn}) and the training phase of the surrogate neural network model.}
\begin{tabular}{lr}
\toprule
\multicolumn{2}{c}{Neural Network Architecture}\\ 
\midrule
Input size & 40 \\
Output size & 40 \\
Number of layers & 4 \\
Number of weights & 9389 \\
\midrule
Type of layer 1 & batch-norm$^{(a)}$\\
\midrule
Type of layer 2 & bi-linear Convolutive \\
Size of layer 2 & $24\times 3 = 72$ \\
Layer 2 convolutive kernel size & 5\\
Activation function & Rectifier  linear unit \\
\midrule
Type of layer 3 & Convolutive \\
Size of layer 3 & 37 \\
Layer 3 convolutive kernel size & 5\\
Activation function  & Rectifier linear unit \\
\midrule
Type of layer 4 & Convolutive \\
Size of layer 4 & 1 \\
Layer 4 convolutive kernel size & 1\\
Layer 4 regularization & L2($10^{-4}$)\\
Activation function  &linear  \\
\midrule
\multicolumn{2}{c}{Training}\\
\midrule

Optimizer & Adagrad$^{(b)}$\\
mini-batch size & 256\\
number of epochs & 20 \\
Forecast lead time $N_\textrm{f}$ & 1 \\
  \bottomrule
  $^{(a)}$\citet{Ioffe2015BatchShift}\\
$^{(b)}$\citet{DuchiJDUCHI2011AdaptiveHazan} 
\end{tabular}
\end{table}

The batch-norm~\citep{Ioffe2015BatchShift} is only used for the input layer, so its main effect is to standardize the input data. The other layers are convolutional layers (CNN) as described in~\citet{Goodfellow-2016-book}. The bi-linear layer is composed of three sub-layers: the outputs of the sub-layers CNN$_{2b}$ and CNN$_{2c}$ are multiplied, and the result is then concatenated to the sub-layer CNN$_{2a}$ to compute the output of the second layer. The third and fourth layers are standard CNNs. The neural network being residual, the output of the fourth layer is added to the input $\mathbf{x}_k$ to compute the output of the neural network.  The number of epochs (20) used for the training is the number of times each element in the training dataset is used by the neural network in each cycle for optimizing the neural network's weights. Therefore if, for example, the algorithm carries out 50 cycles, the neural network performs a total of 1000 optimization epochs ($20\times50$) after the initialization step.

\subsubsection{\label{sec:init}Initialization of the weights}
The weights of the neural network must be initialized before the first cycle (see Fig.~\ref{fig:algo}). Starting with random weights --- as it is usually done in ML training algorithms --- could lead to convergence issues of the two-step algorithm and also would be computationally inefficient. To address these issues, the weights are initialized by training the neural network on a dataset produced by a cubic interpolation (in space and time) of the observations. To account for the lower quality of interpolated data, values of $\mathbf{P}_k^{-1}$ in the loss function in Eq.~(\ref{eq:loss}) are set to 1 where observations are available and zero elsewhere. By doing so, interpolated values of missing data are only used as an input of the neural network but not as a target. After cross-validation, it has been chosen to set initially $N_\textrm{f}=4$ and the number of epochs to 40. These particular high values can be explained by the absence of any prior knowledge of the initial weights for the first training, and the initial neural network being far from a realistic surrogate model. The other parameters of the training phase are given in Tab.~\ref{tab:nn}.

\section{\label{sec:results}Results and discussion}

\subsection{\label{sec:convergence}Convergence of the algorithm}

\begin{figure}
 \centering
\includegraphics[scale=.75]{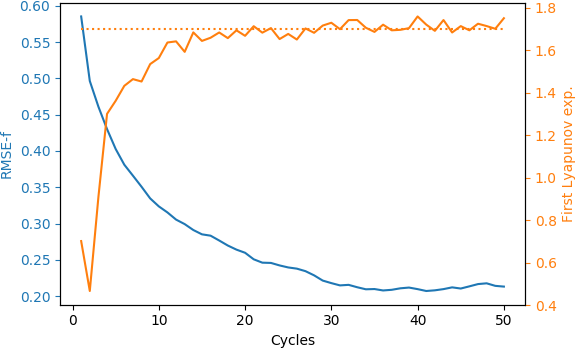}
\caption{\label{fig:convergence} RMSE-f and first Lyapunov exponent as a function of the algorithm cycle. Half of the state-vector is observed, at randomly varying locations. Observational noise standard deviation is $\sigma^\textrm{obs}=1$. The dotted line corresponds to the first Lyapunov exponent of the true model (=1.67)}
\end{figure}

Figure~\ref{fig:convergence} shows the evolution of the RMSE-f and of the first Lyapunov exponent after successive training cycles. The RMSE-f decreases to 0.21 while the first Lyapunov exponent approaches 1.67, which is the target value of the true L96 dynamics.
It can be noticed that none of these quantities is explicitly optimized within a cycle. This suggests that the proposed algorithm is converging toward stable values of two very different targets: RMSE-f is a forecast skill (a sign of accuracy) while the first Lyapunov exponent is a property of the long-term dynamics (a sign of consistency). 

As a remark, note that the initial high value of the Lyapunov exponent may be due to the different parameters $N_\textrm{f}$ and $\mathbf{P}_k$ used in the loss function for the initial training of the neural network (see section~\ref{sec:init}).

\begin{figure}
 \centering
  \includegraphics[scale=.75]{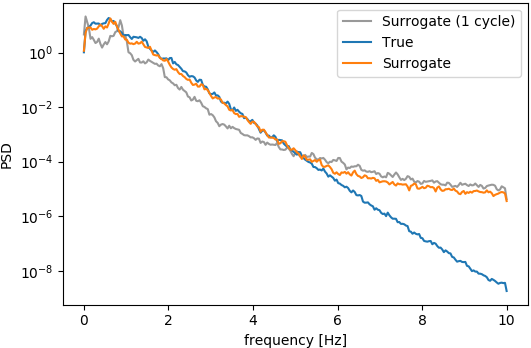}
\caption{\label{fig:psd} Power density spectrum (PSD) of the surrogate model after one cycle (in gray) and after convergence (in orange) compared with the true model (in blue). 50\% of the field was observed with a noise $\sigma^\textrm{obs}=1$.}
\end{figure}

Fig.~\ref{fig:psd} shows the power spectral density of the true L96 model together with the surrogate model after the first cycle (grey line) and after convergence (orange line). After one cycle, some frequencies are favoured (see the peak at $\sim 0.8$Hz) and indicate that the periodic signals are learnt first. 
Results after convergence are discussed in section~\ref{sec:dynamics}.

Given that our primary goal here is to demonstrate the accuracy and convergence of the method rather than optimizing the computational cost, we have intentionally waited for a large number of iterations (50) before stopping the algorithm and retained the surrogate model corresponding to the lowest RMSE-f. A more sophisticated stopping criterion would have reduced the computational burden and would be required in applications in high dimension. 

\subsection{\label{sec:interpolation}Interpolation}
The observations being a sub-sample from the full state vector implies that DA acts as an interpolator and a smoother of the observations, producing a so-called ``analysis''.  Regarding the complete algorithm merging DA and ML, the interpolation of the observations can be viewed as a side-product.
To evaluate the analysis, we compare the RMSE-a with an upper and a lower bound. The former is the RMSE of a field obtained via cubic interpolation in space and time without any use of a dynamical model (which is instead essential in DA); the latter is the RMSE of the analysis obtained using the EnKF-N with the true L96 model instead of the surrogate.

Results are reported in Tab.~\ref{tab:interp}. It can be seen that the use of the surrogate model as an ingredient in the DA leads to a significant error reduction over the cubic interpolation, by a factor of 3 (cf the second and third line in Tab.~\ref{tab:interp}). Furthermore, the RMSE-a is below the observation standard deviation (0.8; recall that $\sigma^\textrm{obs}=1$) and testifies of the healthy functioning of DA. On the other hand, the surrogate model is less accurate than the true model by a factor of 2 (cf the first and second rows in Tab.~\ref{tab:interp}). This means that the data-driven surrogate model does not describe all the dynamics of the underlying process, which should be expected given that the observations are partial and noisy. Recall also that the true model output is hidden from the neural network.

These results suggest that this method could also be used to interpolate spatiotemporal fields in the absence of assumption on the underlying dynamics. More generally, they indicate that the success of the method is coming both from DA and ML, and not from one of the two approaches alone. 

\begin{table}
  \caption{\label{tab:interp}RMSE-a of the complete field estimated with 50\% of observation and a noise  $\sigma^\textrm{obs}=1$}
\begin{tabular}{lr}
\toprule
Method & RMSE-a \\
\midrule
DA with true model & 0.34 \\
DA with surrogate model & 0.80 \\
Cubic interpolation & 2.32 \\
\bottomrule
\end{tabular}
\end{table}

\subsection{\label{sec:dynamics}Emulating the underlying dynamics}
The capabilities of our approach to emulate the underlying dynamics which has generated the observations is assessed by comparing the average value, the power spectral density and the Lyapunov spectrum of the surrogate model with those of the true model.

The average value of the true model is 2.35 whereas one of the surrogate model is 2.30. The surrogate model has a small negative bias due to the underestimation of positive extreme values.

The power spectral density is shown in Fig.~\ref{fig:psd}. The result of the first cycle has already been discussed in section~\ref{sec:convergence}. After convergence, it can be noticed that the surrogate model reproduces the spectrum up to 5 Hz (0.2 time units corresponding to 0.33 Lyapunov time units) but then adds high-frequency noise. The errors are thus larger at higher frequencies and there is no significant improvement throughout the optimization process (see the grey and orange lines after 5 Hz in Fig.~\ref{fig:psd}). The low frequencies are better observed (there are 4 times as many observations for each 2 Hz oscillation as for an 8 Hz oscillation) and better reproduced after the DA step, whereas high-frequencies are not reproduced from observations.

Note that the power spectral density has been computed using a long simulation (1336 Lyapunov time units,  corresponding to 16,000 time steps), which means that the surrogate model is stable enough to compute long-term simulations.

\begin{figure}
 \centering
  \includegraphics[scale=.75]{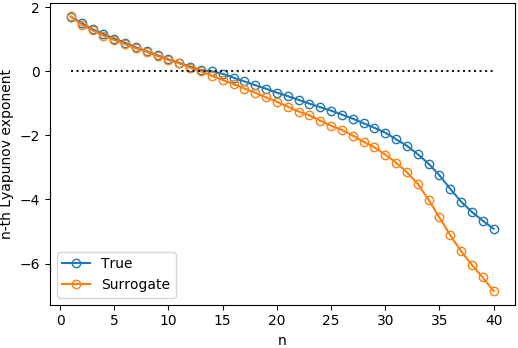}
\caption{\label{fig:lyapu} 40 Lyapunov exponents of the surrogate model (in orange) compared with the true model (in blue). 50\% of the field was observed with a noise $\sigma^\textrm{obs}=1$.}
\end{figure}

Figure~\ref{fig:lyapu} shows the 40 Lyapunov exponents of both models. 
Remarkably, the positive part of the spectrum (the first twelve exponents) of both models are very close to each other. This indicates that the surrogate model possesses an unstable subspace (the time-dependent space where perturbations grow) with the same general properties, size and average growth rate, as the truth. This means that initial errors would grow, on average, in a similar way in the two models that will thus also share the same e-folding time (uniquely determined by the first Lyapunov exponent) and Sinai-Kolmogorov entropy (given by the sum of the positive Lyapunov exponents). The situation is different in the null and negative parts of the spectrum, where the surrogate model possesses smaller (larger in absolute value) exponents. A closer inspection reveals that this shift is mainly due to the difference in the null part of the spectrum.
Recall that the L96 possesses one null exponent and two very small ones; the surrogate model, on the other hand, displays a single null exponent. The difficulty to reproduce the null part of the Lyapunov spectrum has also been put forward by~\citet{Pathak2017UsingData}. While a robust explanation is still missing we argue here that this behaviour can be related to the known slower convergence of the instantaneous null exponent(s) toward their asymptotic values  compared to the positive and negative parts of the spectrum \citep[linear versus exponential; see][]{Bocquet2017DegenerateSubspace}.
It is finally worth noting that the average flow divergence, which is given by the sum of the Lyapunov exponents, is smaller in the surrogate model than in the truth, implying that volumes in its phase space will contract, on average, faster than in the case of the true model (volumes are bound to contract given the dissipative character of the dynamics). 
The average flow divergence also drives the evolution of the probability density function (PDF) of the state in the phase space according to the Liouville equation: the results in Fig.~\ref{fig:lyapu} thus suggest that, once the PDF is interpreted as an error PDF about the system state estimate, the surrogate model is, on average, slightly more predictable in the sense that the PDF support shrinks faster. 

\subsection{Forecast skill}
\begin{figure}
 \centering
\includegraphics[scale=.75]{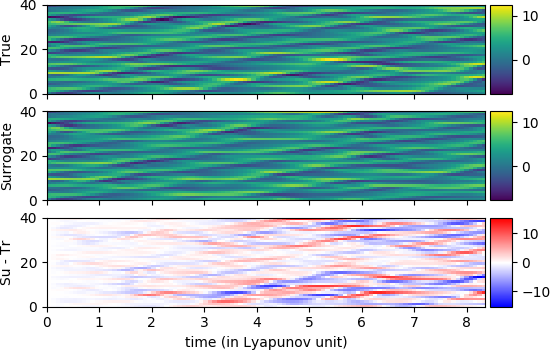}
\caption{\label{fig:simul} Hovm{\o}ller plot of one trajectory of the true model, the surrogate model and their difference given the same initial condition with respect with the lead time (in Lyapunov time). 50\% of the field is observed with a noise level $\sigma^\textrm{obs}=1$.}
\end{figure}

In Fig.~\ref{fig:simul}, we compare one simulation of the true model (top panel) with the surrogate model (middle panel) for about 8 Lyapunov time units. Both simulations have the same initial condition. It can be noticed that the difference (bottom panel) between the true and the surrogate simulation is increasing with respect to time. After 2 Lyapunov time units, the trajectory of the surrogate model diverges significantly from that of the true model and the error saturates after 4-5 Lyapunov time units.

\begin{figure}
 \centering
\includegraphics[scale=.75]{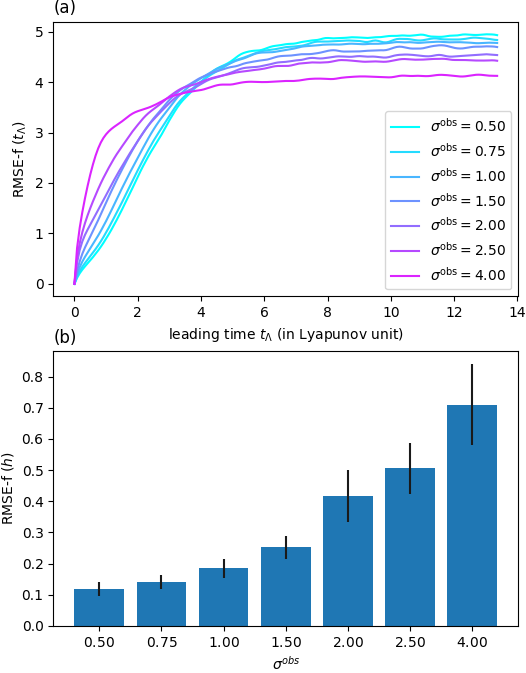}
\caption{\label{fig:rms_noise} (a): RMSE-f of the surrogate model compared with the true model with respect to the leading time for different observational error standard deviation. (b): RMSE-f for the first time step of the surrogate model with respect to the level of noise, vertical black lines represent the standard deviation of the RMSE-f. The surrogate model has been trained on different noise levels ($\sigma^\textrm{obs}$from 0 to 4). The values are averaged over simulations starting from 500 different initial conditions at $t=0$. $50$\% of the field is observed.}
\end{figure}

We quantify the effect of noise on the forecast skill in Fig.~\ref{fig:rms_noise} which shows the RMSE-f of the surrogate model as a function of the forecast lead time (panel (a)) and of the  observational noise level (panel (b)) in the case of a 50\% observed field.
As expected, the forecast skill deteriorates gradually as the observation noise increases. 
It is worth noticing that the asymptotic values of RMSE-f are inversely proportional to the noise level. This is due to the fact that the surrogate model trained with noisy observations underestimates the variance of the forecast (the spatial variability is reduced) and by consequence has a reduced RMSE-f after the predictability horizon. It can also be qualitatively seen in Fig.~\ref{fig:simul} in which it appears that the surrogate model produces less extreme values than the true model.

\begin{figure}
 \centering
  \includegraphics[scale=.75]{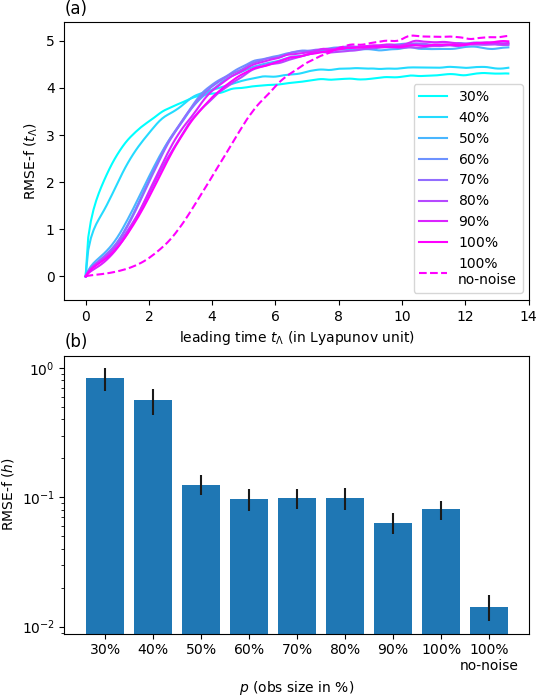}
\caption{\label{fig:rms_nobs} (a): RMSE-f of the surrogate model compared with the true model with respect to the leading time for different observation densities. (b): RMSE-f for the first time step of the surrogate model with respect to the observation density, vertical black lines represent the standard deviation of the RMSE-f. The surrogate model has been trained with increasing observation density (from 30\% to 100\%). The values are averaged over simulations starting from 500 different initial conditions. The noise of the observation is $\sigma^\textrm{obs}=1$ except for the ``100\% no-noise'' value that corresponds to $\sigma^\textrm{obs}=0$. Note that the error bars are relatively small so that their asymetry with respect to the mean due to the use of the log-scale (with an above-to-mean side shorter) is not apparent.}
\end{figure}

In Fig.~\ref{fig:rms_nobs}, we study the sensitivity of the forecast skill to the observation density. Observational noise is fixed to $\sigma^\textrm{obs}=1 $ following the protocol detailed in section~\ref{sec:setup} but the fraction of the observed model domain varies from 30\% to 100\%. 
The RMSE-f of the surrogate model trained with complete (i.e. 100\%) perfect (i.e. $\sigma^\textrm{obs}=0$) observations is also displayed for reference. Since the observations are ``perfect'', DA has not been carried out in this case. The RMSE-f at $t_0+h$ (see the right-most bar in panel (b)) is $0.014$ which is close to the value in~\citet{fablet2018} despite the fact that, in our case, the model is not identifiable. 
If the field is observed with less than 50\% coverage, the skill is significantly degraded. At the same time, when more than 50\% of observations are available, there is no obvious improvement either. These results suggest that, to a certain extent,  the DA phase of the algorithm is very efficient in producing accurate analyses from a sparsely observed field, which is also confirmed by the scores in Tab.~\ref{tab:interp}. The threshold of 50\% of observations is likely to be a value specific to the L96 model. Note also that the same set of parameters has been used for all the experiments presented (architecture of the neural net, additive stochastic model noise, ...) whereas specific tuning for each experimental setup would be desirable and could lead to improved performances. If 100\% of the field is observed with an error corresponding to $\sigma^\textrm{obs}=1$, there is still a significant degradation with respect to the perfect case $\sigma^\textrm{obs}=0$. This can be due to several factors: first, there is a severe loss of information when noise is added; then, the DA step (which is not performed in the perfect observation case) does not provide a perfect estimate of the state (even in the case of a perfect model); finally, the algorithm itself is also a source of error since it could converge toward a local minimum.

\subsection{\label{sec:tuning}Sensitivity to hyperparameters}
We study here the effect of two control parameters:the size of the training set $K$, and the model noise by varying its standard deviation $\sigma^\textrm{m}$ defined in Eq.~(\ref{eq:forecast}).
In principle, $\sigma^\textrm{m}$ should have been tuned for each cycle and observational time series considered. Nevertheless, we choose a common tuning to facilitate the comparison among  cases. In the following, we evaluate the effect of these parameters on the reference setup ($p=20$  and $\sigma^\textrm{obs}=1$) using two metrics: the RMSE-f defined in Eq.~(\ref{eq:rmse-f}) and the RMSE between the first 12 exponents of the Lyapunov spectrum (corresponding to the positive Lyapunov exponents in the true model):
\begin{equation}
    \text{RMSE-Lyapunov} = \sqrt{\sum_{n=1}^{12} \left(\Lambda^{\mathcal{G}_\textrm{L96}}_n - \Lambda^{\mathcal{G}_\mathbf{W}}_n\right)^2},
\end{equation}
where $\Lambda^{\mathcal{G}_\textrm{L96}}_n$ (resp. $\Lambda_n^{\mathcal{G}_\mathbf{W}}$) is the $n$-th Lyapunov exponent for the true model (resp. surrogate model). While the
RMSE-f estimates the forecast skill of the surrogate model, the RMSE-Lyapunov assesses the algorithm capability to reconstruct the long-term chaotic dynamics.
The sensitivity to the forecast lead time $N_\textrm{f}$ included in the loss, as defined in Eq.~(\ref{eq:loss}), was also tested. For all the metrics considered, the best results were obtained for $N_\textrm{f}=1$ (not shown).



\subsubsection{\label{sec:tuning-K}Size of the training set}
\begin{figure}
 \centering
\includegraphics[scale=.75]{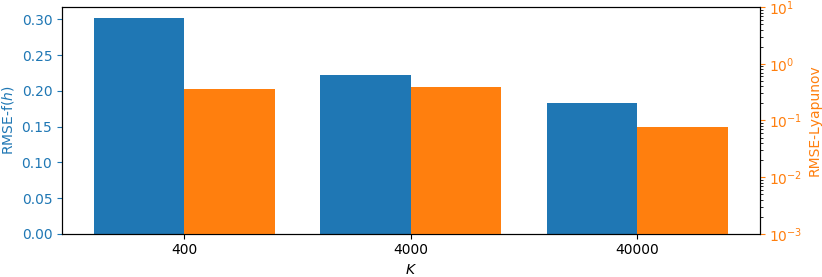}
\caption{\label{fig:ntrain} RMSE-f (blue) and RMSE-Lyapunov (orange) with respect to the size of the training set $K$. 50\% of the field was observed with a noise $\sigma^\textrm{obs}=1$.}
\end{figure}

Figure~\ref{fig:ntrain} shows the value of RMSE-f and RMSE-Lyapunov for three different training test size $K \in \{4\times10^2,4\times10^3,4\times10^4\}$. Note that $K=4\times10^4$ is the value used for all the other experiments in this paper. As expected, the error of the algorithm for both metrics decreases as the size of the training set increases. Note that the amplitude of the scale used in this figure is small compared with the other sensitivity experiment (Fig.~\ref{fig:mnoise}). The result shows that it is possible to produce a fair surrogate model using this algorithm with smaller datasets.

It is worth noticing that making the algorithm to converge is more challenging with $K<4\times10^4$. This is likely due to the fact that the neural network can overfit the dataset during the training phase. In this particular experiment, the number of epochs for each cycle has been reduced to 5 (instead of 20 in the other experiments as stated in Tab.~\ref{tab:nn}) to mitigate overfitting. We do also recognise the possible model-dependent character of these results so that another dynamics than the L96 model could show a different sensitivity to the size of the dataset.

\subsubsection{Model error}
\begin{figure}
 \centering
\includegraphics[scale=.75]{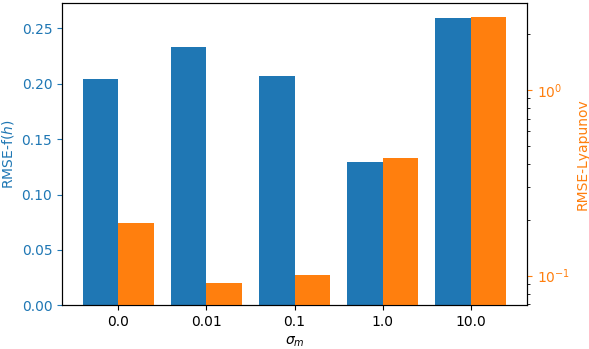}
\caption{\label{fig:mnoise} RMSE-f (blue) and RMSE-Lyapunov (orange) with respect to the standard deviation of the model error $\sigma^\textrm{m}$, Eq. ~(\ref{eq:forecast}). 50\% of the field is observed with a noise $\sigma^\textrm{obs}=1$.}
\end{figure}

Figure \ref{fig:mnoise} shows the sensitivity of the same skill scores with respect to the model noise specified in the DA process in Eq.~(\ref{eq:forecast}). 
Regarding the forecast skill, the optimal value seems to be $\sigma^\textrm{m} \approx 1.0$ whereas the reconstruction of the long-term dynamic favours a much smaller $\sigma^\textrm{m} \approx 0.01$. It means that this parameter can be adjusted to the purpose of the surrogate model: either yielding accurate forecast or reconstructing consistently the long-term dynamics. In our case, we have selected a compromise by setting $\sigma^\textrm{m} = 0.1$.

We tentatively explain the reason for this trade-off between forecasting skill and consistent reconstruction as follows.
The larger $\sigma^\textrm{m}$ is, the larger the spread of the ensemble in the DA step and the more efficient is the DA. On the other hand, the DA updates interfere with the calculation of the Lyapunov exponents. With a small model error $\sigma^\textrm{m}$, such jumps in the time series are much smaller. 

A perspective of this work, which is outside the scope of this paper, would be to propose some methodology to estimate the model error statistics~\citep[see e.g.][]{Pulido2018StochasticMethods}. Note finally that, up to a certain extent, the implicit estimation of the multiplicative inflation in the EnKF-N can compensate for some model error even if it was not initially targeted for that~\citep{Raanes2019AdaptiveMixtures}.

Another ongoing direction of the extension of the current work could be to estimate stochastic model errors. For example, generative models (generative adversarial networks or variational auto-encoders) could be used to tune stochastic parametrisation schemes.
\section{Conclusions}  
\label{sec:conclusion}
We have proposed a methodology to build a data-driven surrogate model from partial and noisy observations. The method is based on an iterative algorithm. At each iteration, a data assimilation step interpolates and de-noises the observations, then alternates with a machine learning step that learns the underlying dynamics of the DA analysis.

The approach has been illustrated on the spatially extended (one dimensional) 40-variable Lorenz chaotic model~\citep{Lorenz1998OptimalModel} and evaluated by three main skill scores. First, we have considered the ability to interpolate and de-noise the observational field. In our reference experiment (50\% of the field observed, with a standard deviation of noise set to 1), it has been shown that the proposed method was able to reconstruct the complete field with a smaller error than the observational error. Second, we have evaluated the dynamical features of the surrogate model compared with the true L96 model. We have shown that the more energetic time scales (frequency lower than 5 Hz corresponding to $t>0.33$ in Lyapunov time unit), and the positive Lyapunov exponents are well represented by the surrogate model. Finally, we have evaluated the forecast skill of the model, achieving accurate forecasts out to $2$ Lyapunov time units (twice the typical "memory" of the system).

Sensitivity experiments to the level of observation noise and to the density of available observations have been conducted.  The method is very sensitive to the level of noise, as expected, and the forecast skill gets worse with an increasing noise level. If less than 50\% of the field is observed, the results also deteriorate but as long as the fraction of observations exceeds 50\% there is no significant difference in the skill of the surrogate model. In the case of a fully observed field without noise, the surrogate model reproduces the skill found in similar machine learning configurations although not as accurate as in~\citet{Bocquet2019DataModels} in the case of an identifiable model.

It has also been shown that by tuning certain parameters of the algorithm (number of forecast steps of the neural network and standard deviation of the model noise in data assimilation), it was possible to favour the forecast skill over the long-term dynamics reconstruction or vice-versa.

One drawback of the method is the computation cost. The algorithm needs to apply one complete data assimilation procedure (equivalent to 30 forward model runs to propagate the ensemble members) and one neural network training (equivalent to 20 forward model runs) for each cycle until convergence. In this work, we have not optimized the computational cost of the algorithm, but there are several avenues for improvement of that aspect. For a model defined on larger domains, the cost is expected to be dominated by the model integration time, which scales linearly with the domain size for a convolutional neural network-based model. We could, for example, set an optimal stopping criterion, or leverage from concurrent computing (the neural network could start training before the end of the data assimilation, using the portion of the assimilation run which is already analysed). We could also more systematically benefit from accelerations enabled by parallel computing, by using for instance graphical processor units (GPU), given that deep-learning libraries offer an easy-to-use GPU version of the algorithms.

Arguably, the success of the method relies at least partially on the autonomous character of the L96 model. This makes sure that learning from past analogues is valuable for forecasting purposes. Nevertheless, we foresee that the approach can be further extended to non-autonomous large scale systems by considering that the system variations are slow or by including the external forcing as an input of the neural network.  A major feature of the proposed algorithm is its extreme flexibility. It is possible to plug any data assimilation scheme and machine learning schemes independently. For example, from an existing data assimilation system, it could be feasible, without much change in the existing system, to apply this algorithm to estimate the surrogate model of a non-represented dynamical process in the original numerical model (bias, parameter, ...) or to replace one particular heuristic sub-model by a trained neural network. The latter would result in a hybrid numerical model, including trained neural networks, that honours the fundamental equations of the motions and only relies on machine learning for a  data-driven sub-model. In term, one could envision that model tuning and data assimilation could be carried out in the same framework.

\appendix

\section*{Competing interests}
{The authors declare they have no conflict of interest} 

\section*{Acknowledgements}
JB, AC and LB have been funded by the project REDDA (\#250711) of the Norwegian Research Council. AC has also been funded by the Trond Mohn Foundation under the project number BFS2018TMT0. CEREA and LOCEAN are members of Institut Pierre–Simon Laplace (IPSL). AC has also been supported by the Natural Environment Research Council (Agreement PR140015 between NERC and the National Centre for Earth Observation). 

The authors thank Peter Dueben for his useful suggestions reviewing a previous version of this work and thank the two reviewers for their comments that helped to improve the quality of this article.

\printcredits

\bibliographystyle{cas-model2-names}
\bibliography{references.bib}
\newpage

\bio{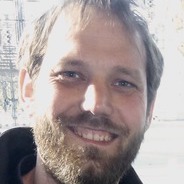}
Julien Brajard is associate professor at Sorbonne University (Paris, France) since 2009 and researcher at NERSC (Nansen Environmental and Remote Sensing Center) since 2018. He works on machine learning approaches applied to environmental and climate science. He has published 16 peer-reviewed papers.
\vspace{4em}
\endbio

\bio{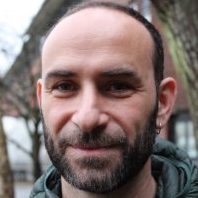}
Alberto Carrassi is professor of data assimilation at University of Reading (UK) and adjunct at University of Utrecht (NL). He works at the crossroad between dynamical systems and data assimilation for atmospheric and ocean science and has published 40 peer-reviewed articles and 2 chapters in collective volumes.
\vspace{4em}
\endbio

\bio{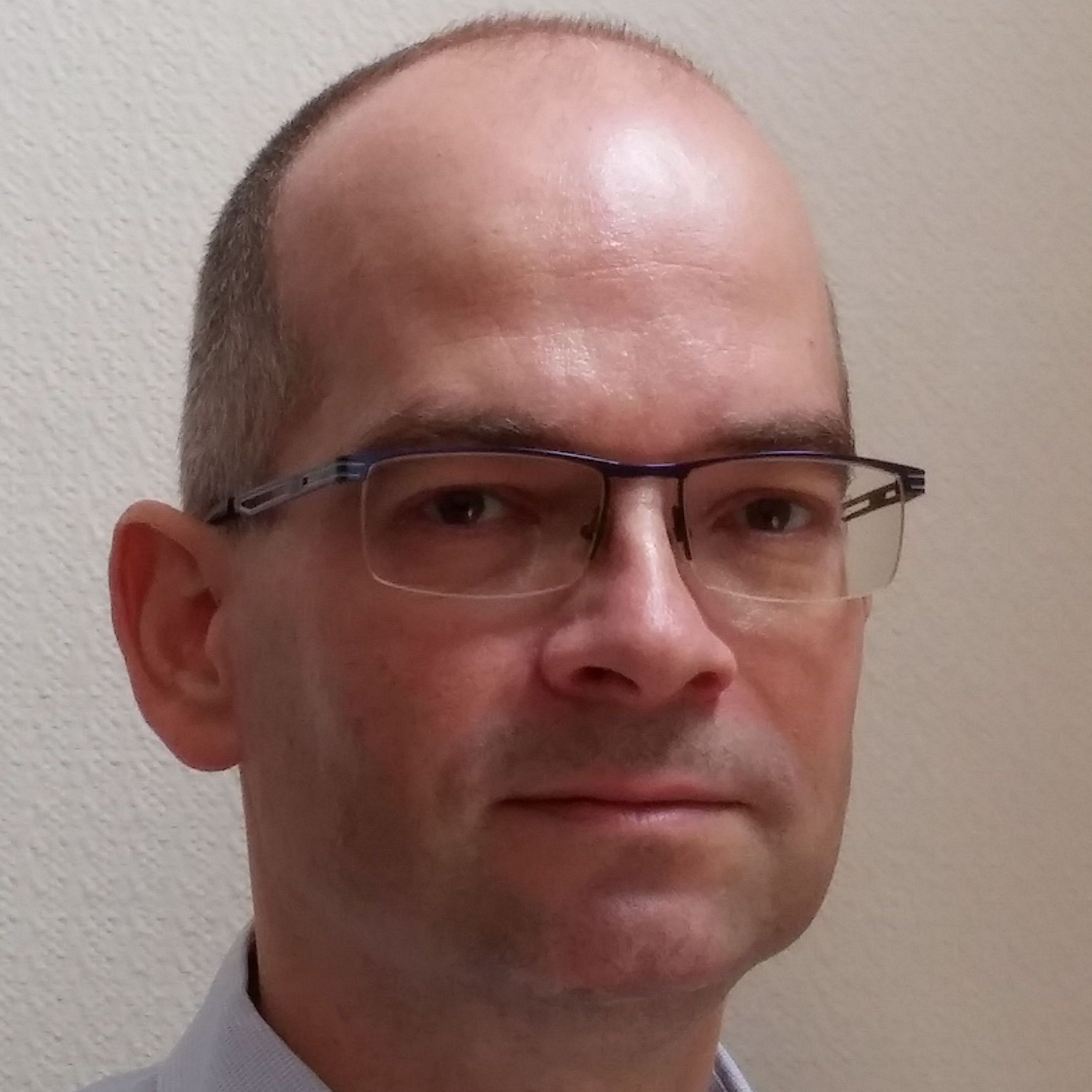}
Marc Bocquet is Professor at Ecole des Ponts ParisTech (France) and deputy director of CEREA laboratory. He works on the methods of data assimilation, environmental statistics and machine learning,
with applications to dynamical systems, atmospheric chemistry and transport, and meteorology.
He has published 103 peer-reviewed papers.
\vspace{4em}
\endbio

\bio{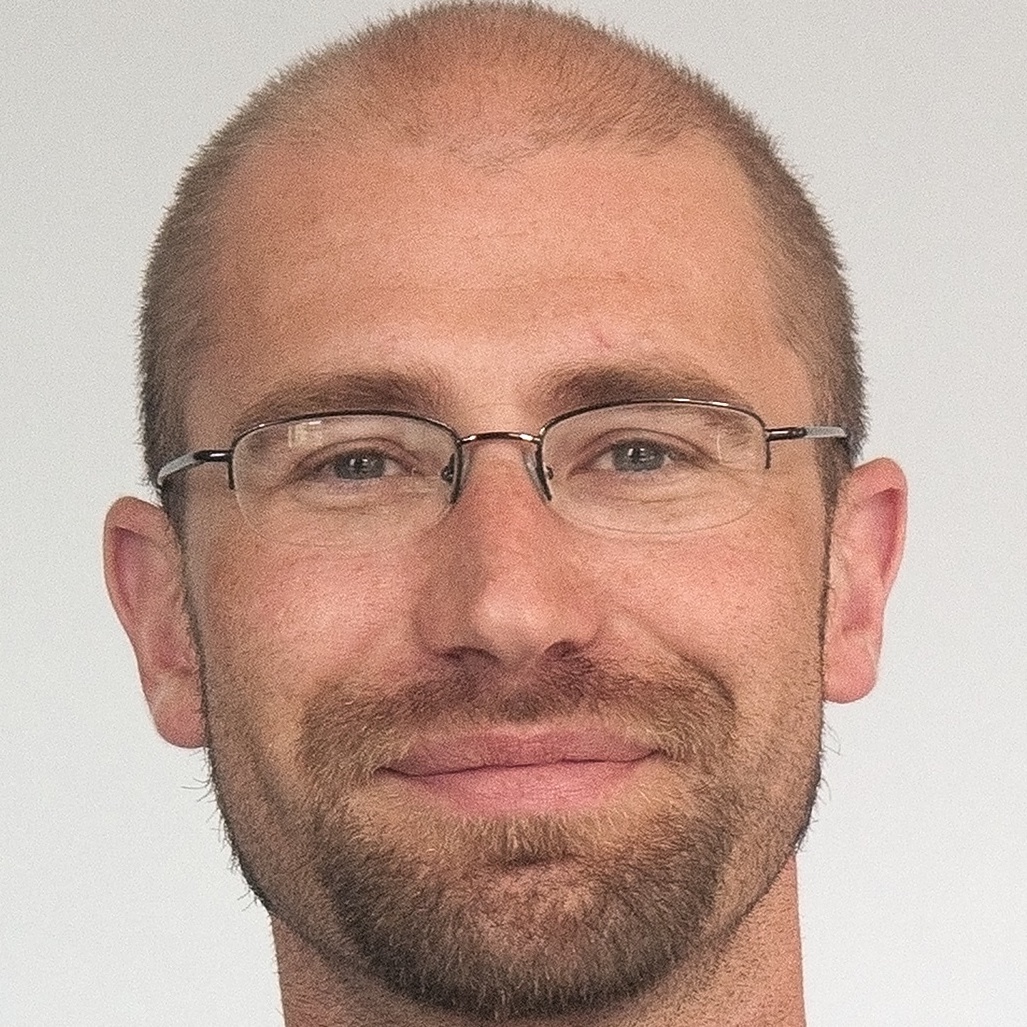}
Laurent Bertino is a Research Leader at NERSC with a background in Geostatistics and data assimilation. He works on ensemble-based data assimilation techniques with applications to oceanography and sea ice forecasting in the Arctic. He has published 84 peer-reviewed papers.
\vspace{4em}
\endbio

\end{document}